\documentclass{article}

\usepackage{arxiv}

\usepackage[utf8]{inputenc} 
\usepackage[T1]{fontenc}    
\usepackage{hyperref}       
\usepackage{url}            
\usepackage{booktabs}       
\usepackage{amsfonts}       
\usepackage{nicefrac}       
\usepackage{microtype}      
\usepackage{graphicx}
\usepackage{natbib}
\usepackage{doi}

\usepackage{algorithm}
\usepackage{algorithmic}
\usepackage{amsmath}

\usepackage{amsmath,amsfonts,bm}

\def\eqref#1{equation~\ref{#1}}

\def\1{\bm{1}}

\DeclareMathAlphabet{\mathsfit}{\encodingdefault}{\sfdefault}{m}{sl}
\SetMathAlphabet{\mathsfit}{bold}{\encodingdefault}{\sfdefault}{bx}{n}

\newcommand{\E}{\mathbb{E}}

\usepackage{enumitem}
\usepackage[title]{appendix}

\newcommand{\mi}{mini-ImageNet~}
\newcommand{\fr}{federated reconnaissance}
\newcommand{\frb}{Federated Reconnaissance Benchmark}

\title{{\bf Federated Reconnaissance:\\
           Efficient, Distributed, Class-Incremental Learning}}

\date{} 					

\author{Sean M.~Hendryx \thanks{Work partially completed at Stanford University.} \\
	School of Information\\
	University of Arizona\\
	\texttt{seanmhendryx@arizona.edu} \\
	\And
	Dharma Raj KC \\
	Department of Computer Science\\
	University of Arizona\\
	\texttt{kcdharma@email.arizona.edu}  \\
    \AND
    Bradley Walls \\
    Aret\'e Associates\\
    \texttt{bwalls@arete.com} \\
    \And
    Clayton T. Morrison \\
    School of Information\\
    University of Arizona\\
    \texttt{claytonm@arizona.edu}
}

\renewcommand{\headeright}{}
\renewcommand{\undertitle}{}
\renewcommand{\shorttitle}{Federated Reconnaissance}

\hypersetup{
pdftitle={Federated Reconnaissance: Efficient, Distributed, Class-Incremental Learning},
pdfsubject={Machine Learning},
pdfauthor={Sean M.~Hendryx},
pdfkeywords={Machine Learning, Continual Learning, Distributed Learning, Meta-Learning},
}

\begin{document}
\maketitle

\begin{abstract}
	We describe federated reconnaissance, a class of learning problems in which distributed clients learn new concepts independently and communicate that knowledge efficiently. In particular, we propose an evaluation framework and methodological baseline for a system in which each client is expected to learn a growing set of classes and communicate knowledge of those classes efficiently with other clients, such that, after knowledge merging, the clients should be able to accurately discriminate between classes in the superset of classes observed by the set of clients. We compare a range of learning algorithms for this problem and find that prototypical networks are a strong approach in that they are robust to catastrophic forgetting while incorporating new information efficiently. Furthermore, we show that the online averaging of prototype vectors is effective for client model merging and requires only a small amount of communication overhead, memory, and update time per class with no gradient-based learning or hyperparameter tuning. Additionally, to put our results in context, we find that a simple, prototypical network with four convolutional layers significantly outperforms complex, state of the art continual learning algorithms, increasing the accuracy by over 22\% after learning 600 Omniglot classes and over 33\% after learning 20 mini-ImageNet classes incrementally. These results have important implications for federated reconnaissance and continual learning more generally by demonstrating that communicating feature vectors is an efficient, robust, and effective means for distributed, continual learning.
\end{abstract}


\section{Introduction}
\label{intro}

In this work, we present {\em federated reconnaissance}, a new a class of learning problems in which distributed models should be able to learn new concepts independently and share that knowledge efficiently. Typically in federated learning, a single static set of classes is learned by each client~\citep{mcmahan2017communication}. In contrast, federated reconnaissance requires that each client can individually learn a growing set of classes and communicate knowledge of previously observed and new classes efficiently with other clients. This communication about learned classes permits {\em merging} the knowledge from the clients; the resulting merged model is then expected to support the superset of the classes each client has been exposed to. The merged model can then be deployed back out to the clients for further learning. In practice, a client in a distributed system may only see a small number of examples for a new class. This problem therefore stands at the intersection of large bodies of work on continual, meta-, and federated learning.
Examples of this problem include mobile phone applications in which the service should be able to both learn to identify instances of a new class added by a user and transfer that classification ability to other client devices.
In this paper, we outline related work, formalize the federated reconnaissance problem statement, introduce a benchmark by adapting \mi~\citep{vinyals2016matching}, compare a range of learning algorithms and neural network architectures, and find that a simple algorithm adapting prototypical networks~\citep{snell2017prototypical} is a strong baseline for both single client continual learning and federated reconnaissance. Our code and pretrained models are available at: \url{https://github.com/ml4ai/fed-recon}.

\subsection{Prior Work}

Continual learning of new concepts is an open and long-standing problem in machine learning and artificial intelligence with no semblance of a unified solution~\citep{thrun1995lifelong, lopez2017gradient, shin2017continual, zenke2017continual, van2019three, farajtabar2020orthogonal}. While deep neural networks have proven to be incredibly effective in a wide range of tasks, the available methods for continuously integrating new information whilst remembering previously learned concepts suffer from being either compute inefficient (in the case of algorithms that retrain on a cache of examples~\citep{rebuffi2017icarl} or on generated examples~\citep{van2020brain}) or lacking in expressivity and accuracy (in the case of regularization-based methods~\citep{kirkpatrick2017overcoming}). Much earlier work on continual learning (see \citet{van2019three} and their citations) focused on learning classes or tasks sequentially from scratch. While this is an interesting problem setting for studying neural memory, it can be impractical and unnecessary for deployment to production systems. In more recent work on continual learning, the authors in~\citet{javed2019meta, prabhu2018prototypical, beaulieu2020learning} change the continual learning problem setup
by assuming access to a set of pretraining data, often referred to as a set of meta-training tasks, and find that such pretraining can benefit later continual learning.
Access to a pretraining dataset is a reasonable assumption for production systems and enables the development of algorithms that can first learn invariances that can later be exploited when learning new classes online.
In this work, we assume access to a set of pretraining data and explore algorithms that allow for the efficient and accurate learning of new classes sequentially.

In contrast to the common variants of continual learning, federated learning iteratively trains a common model under the direction of a central server on data that is decentralized across many devices, such as mobile phones. Federated learning can reduce communication overhead and privacy concerns by removing the need to send the raw source data back to a server for traditional, centralized machine learning~\citep{mcmahan2017communication, kairouz2019advances}. The goal in federated learning has traditionally been to learn a shared parameterization from decentralized data, but not necessarily to learn \textit{new} concepts or classes online on different clients while preserving that discriminative information
 when communicating between client and the server. Recent work~\citep{yoon2020federated} discusses federated continual learning yet assumes that each client learns distinct tasks. While they address federated continual learning directly, the authors do not consider the direct sharing of knowledge of classes. Instead they assume that each client is learning its own task and they focus on reducing interference when transferring the network's parameters, but not on merging explicit knowledge of classes that have been seen by the clients. In many cases, it would instead be useful if the server model could learn a single task with a growing set of classes by incorporating knowledge of classes learned on client devices. Such a unified, class-incremental learning model could be valuable to users of mobile devices, intelligence operations, robotics, or any situation in which new concepts should be learned and shared between distributed clients when efficient communication, privacy, and/or fast learning are paramount. Because this work builds on the motivation of federated learning but with the explicit goal of ascertaining knowledge of new concepts that can be efficiently communicated and reused, we call this problem federated reconnaissance.

\subsection{Contributions}

Federated reconnaissance poses a unique set of challenges. An effective federated reconnaissance system must tackle efficient in situ learning of new classes and knowledge-preserving transfer. To these ends, we systematically study different approaches to solving this problem including running stochastic gradient descent (SGD) on new data as it appears as an empirical lower bound, an iCaRL~\citep{rebuffi2017icarl} algorithm adapted for federated reconnaissance, an extension of prototypical networks for distributed, continual learning, and, finally, SGD on the joint distribution of all training data from all clients as an empirical upper bound.

We posit that, when a pretraining dataset is available, prototypical networks~\citep{snell2017prototypical} are a strong baseline for federated reconnaissance due to:
\begin{enumerate}
 \item The ability to compress concepts into relatively small vectors known as prototypes, enabling efficient communication,
 \item Robustness to catastrophic forgetting when learning on non-IID data\footnote{In distributed online learning, the data is not guaranteed to be IID over time nor across clients since each client may observe a local distribution of potentially correlated examples~\citep{zhang2020personalized}.}, and
 \item Enabling fast knowledge transfer as no gradient-based learning or hyperparameter tuning are required during model merging.
\end{enumerate}
To test this claim, we present two simple algorithms for using prototypical networks for federated reconnaissance and evaluate them on the federated reconnaissance \mi benchmark, showing that they outperform the lower bound and iCaRL models handily in both accuracy and computational complexity. We go on further to present pretraining methods that increase the accuracy of prototypical networks on the federated reconnaissance \mi benchmark. Additionally, to put federated prototypical networks into the context of previous work, we show that they substantially outperform recent state of the art works on few-shot continual learning from~\citet{javed2019meta, beaulieu2020learning}. It is also worth noting that prototypical networks have stronger privacy protection than existing class-incremental replay-based learning algorithms such as iCaRL since the transfer of raw examples is avoided and as long as a minimum set of examples per class are averaged on each client. We leave the empirical and theoretical work of differential privacy for federated prototypical networks to future work.

Finally, we hypothesize the existence of a distributed learning phenomena, \textit{collective ascent}, which occurs when the accuracy of a group of clients increases due to knowledge sharing. Collective ascent is trivial to identify in centralized model training with distributed example gathering as it only amounts to greater data collection. In contrast, collective ascent is non-trivial to produce during federated reconnaissance when compute or bandwidth are limited and
centralized retraining on all observed examples is not an option because collective ascent entails positive forward \textit{and} backward transfer\footnote{See~\citet{riemer2018learning} for a discussion of forward and backward transfer and
interference in the context of continual learning.} during distributed continual learning.
We demonstrate collective ascent on the \fr~\mi benchmark, highlighting, in concert with our other results, that the simple procedure of sharing feature vectors representing shared concepts is an important avenue for federated reconnaissance and continual learning more generally.

\section{Federated Reconnaissance Problem Statement}

\subsection{Desiderata}

Federated reconnaissance requires continual learning on each client device, efficient communication, and knowledge merging.
Inspired by applications to learning new classes across a large number of distributed client devices, we define the following desiderata of a federated reconnaissance learning system:
\begin{enumerate}
	\item Each client model should be able to learn new classes in situ from only a few examples and be able to improve accuracy as more examples become available. 
   \footnote{In this work, we assume an in situ source of supervision and identification of new classes. We follow and reproduce the single client continual learning benchmarks of ~\citep{javed2019meta, beaulieu2020learning} in which the model only observes a small number (e.g. $\le$ 30) of examples to learn from at meta-test time.}
	\item After learning new classes, each model should not forget previously seen classes. I.e., the model should not suffer from catastrophic forgetting.
	\item To reduce communication costs and to enable distributed learning when bandwidth is limited, a federated reconnaissance system should be able to compress information before transfer.
	 \item Finally, to avoid costly retraining on all data on the central server each time a new class is encountered by a client, the federated reconnaissance system should be able to merge knowledge of new classes learned by distributed client models quickly.
\end{enumerate}
The specific requirements of a real world implementation of federated reconnaissance will of course determine the details and relative importance of each desideratum.

\subsection{Problem Definition}
Formally, federated reconnaissance consists of a set of clients $\mathbb{C} := \{c_i ~|~ i \in 1 ... C\}$ that each have an exposure history to a growing set of classes $\mathbb{M}_{i, t} := \{p(y = j|x)~|~j \in 1 ... M_{i}\}$ where $C$ indicates the total number of clients, $M_i$ indicates the number of classes that client $C_i$ should be able to discriminate between, and a class is represented as a probability $p(y=j|x)$ of label $j$ given an input $x$. The central server is tasked with merging the clients' knowledge of the superset of classes $\mathbb{M}_{t} = \bigcup_{i=1}^{C} \mathbb{M}_{i,t}$
and deploying an updated model that supports $\mathbb{M}_{t}$ back out to $\mathbb{C}$. A client $C_i$ can be exposed to a new class by training directly on a set of labeled examples $\{(x, y) ~|~ (x, y) \in X_j \times Y_j \}$
 or via communication of compressed knowledge that allows the client to approximate $p(y=j |x)$.

\begin{figure*}
	\centering
	\includegraphics[width=0.9\linewidth]{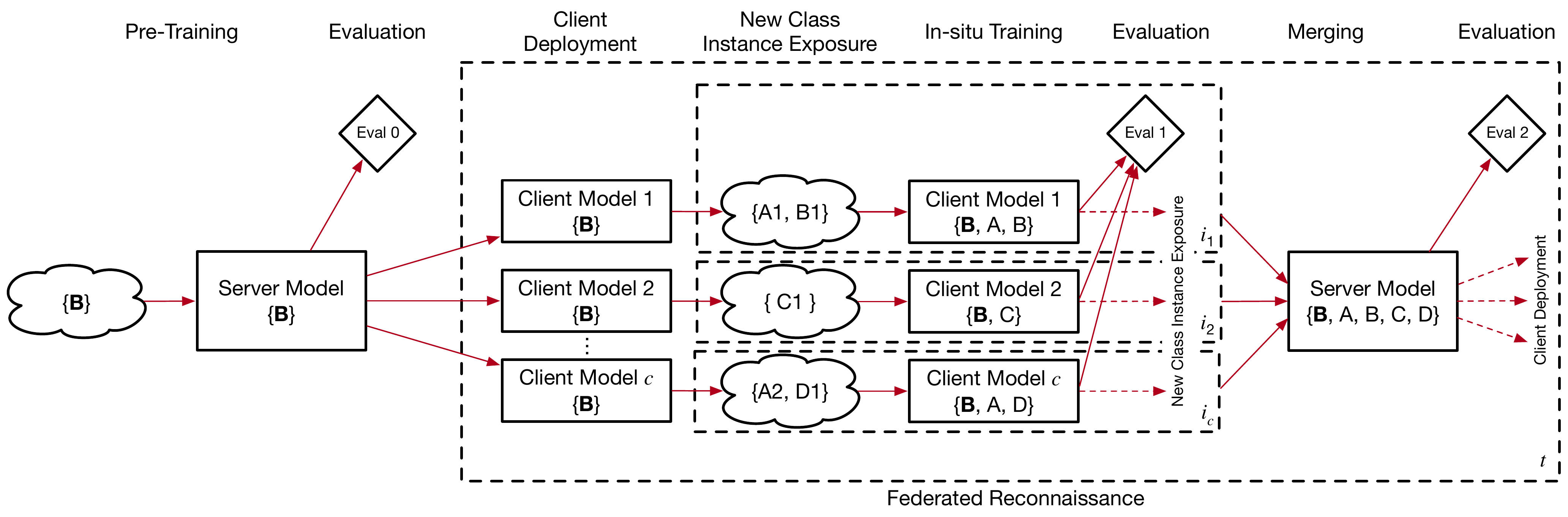}
	\caption{In the above diagram we show the evaluation framework of a federated reconnaissance system. A base model is pretrained on a set of base classes $\mathbb{B}$ and evaluated on held out examples of the classes in $\mathbb{B}$. This base model is then deployed to clients $1..h$. Each client is then trained via local supervision on instances of new or previously seen classes. After each client learns new classes, it is evaluated on the superset of previous and new classes. The clients then communicate their knowledge of new classes back to a central server which in turn deploys the merged knowledge back to clients. Finally, the model is evaluated on all classes in the benchmark.}
	\label{fig:fedreconeval}
\end{figure*}

 An effective federated reconnaissance learning \textit{system} entails accurate prediction of $p(\hat{y}=j|x)$ in expectation over clients in $\mathbb{C}$ regardless of whether or not each individual client learned class $j$ directly from labeled examples or vicariously via communication of compressed knowledge of $j$ from another client. This brings us to the distributed objective function of the federated reconnaissance learning system at any point in time, which is the average loss across clients:  
 \begin{equation}
 	\label{eq:single-outer-step-objective}
 	\mathcal{L}_t = \frac{1}{|\mathbb{C}|} \sum_{i=1}^{|\mathbb{C}|}  \frac{1}{J_i} \sum_{j=1}^{J_i} \frac{1}{K_{i,j}} \sum_{k=1}^{K_{i,j}} H(\hat{y}_{i,j,k}, y_{i,j,k})  
 \end{equation}

 where $J_i$ is the total number of classes that client $C_i$ has seen in its exposure history, $K_{i,j}$ is the number of examples on client $C_i$ for class $j$, $H$ is the cross entropy between the predicted class $\hat{y}_{i,j,k}$ and the labeled class $y_{i,j,k}$ for example $k$. This loss describes the expected loss over distributed clients as illustrated as Eval. 1 in Figure \ref{fig:fedreconeval}. To simplify notation, we assume a fixed number of clients throughout deployment, though the extension to a variable number of clients over time is straightforward.

 For concise terminology, we define a \textit{mission} as an iteration of sending clients out, having them collect and learn from data in situ and in parallel, and then finally communicating their findings back to a central server. At each mission $t$ a few-shot dataset $\mathcal{D}_{t}$ of examples from a set of classes is produced by the environment.
 Federated reconnaissance comprises both problems in which clients begin learning from scratch without any knowledge of $p(X, Y)$ and those in which data from a subset of \textit{base} classes $\mathbb{B}$ is available for pretraining. The set $\mathbb{B}$ is similar to the meta-training set in meta-learning~\citep{finn2017model}, though unlike the typical setup in few-shot low-way meta-learning, we now expect the clients to learn a growing superset of classes which includes both the \textit{base} classes and \textit{field} classes which are learned after a centralized pretraining procedure on $\mathbb{B}$. Access to a dataset representing $\mathbb{B}$ tends to be a reasonable assumption in practice as some number of pretraining classes can usually be measured before a federated reconnaissance system is deployed.\footnote{Further, looking at human intelligence, we find many examples, such as with parenting, education, and training programs, in which a pretraining phase exists before agents go out to learn and communicate new concepts online.}

We define the expected loss directly after model merging by taking the expectation over classes in $\mathbb{M}_t$:
\begin{equation}
	\label{eq:cross-entropy-eval-2}
	\mathcal{L}_t = \frac{1}{|\mathbb{M}_{t}|} \sum_{j=1}^{|\mathbb{M}_{t}|} \frac{1}{K_{i,j}} \sum_{k=1}^{K_{i,j}} H(\hat{y}_{i,j,k}, y_{i,j,k})
\end{equation}

This loss can be discretized into an accuracy metric and its evaluation is illustrated in Eval 2 in Figure \ref{fig:fed-recon-eval}.
\begin{equation}
	\label{eq:acc-eval-2}
	acc_{t} = \frac{1}{|\mathbb{M}_{t}|} \sum_{j=1}^{|\mathbb{M}_{t}|} \frac{1}{K_{i,j}} \sum_{k=1}^{K_{i,j}} \left[ \hat{y}_{i,j,k} = y_{i,j,k}\right]
\end{equation}

At each time step $t$, each client model is first presented with a set of labeled examples of new classes and then evaluated on held out examples from the superset of the new training classes and all classes in it's exposure history. This evaluation step is represented as evaluation diamond 0 in Figure \ref{fig:fedreconeval}. Following in situ, on-client learning, the clients send information back to the server, either communicating information of new classes or updating the information of previously seen classes and the server merges the information from multiple clients together. After this model merging step, we evaluate accuracy on the superset of base and field classes that all clients have seen thus far on held out examples (evaluation diamond 2 in Figure \ref{fig:fedreconeval}). As shown in Figure \ref{fig:fedreconeval}, the process of local client learning and communicating knowledge back to the server is repeated iteratively.
Therefore, we want to minimize the expected loss in equation \ref{eq:single-outer-step-objective} over some horizon of missions $\{t \in \mathbb{N} | t \le T\}$:
\begin{equation}
	\label{eq:expectation-of-objective}
	\min\mathop{\E}_{t \in 1 \dots T} \left[ \mathcal{L}_t \right]
\end{equation}
or alternatively, after the agents have learned for some fixed number of missions:
\begin{equation}
	\label{eq:objective}
	\min \mathcal{L}_{t=T}
\end{equation}
For simplicity of exposition and to highlight the unique challenges of continual distributed learning, our reported tabular results on the \mi benchmark evaluate accuracy at the end of all missions according to \ref{eq:objective}.

 It is worth noting that because the single client in situ update is a non-IID class incremental learning problem, federated reconnaissance generalizes class incremental learning to multiple clients. Therefore in this work, we evaluate a range of learning algorithms and backbones on both the federated reconnaissance problem and prototypical networks on the single client, few shot continual learning benchmark proposed by~\citet{javed2019meta}.

\section{Methods}

\subsection{Learning Algorithms}
For federated reconnaissance, we compare the following four approaches to distributed, continual learning:
\begin{enumerate}
\item \textbf{Lower Baseline}: To demonstrate the learning dynamics and the worst case effects of catastrophic forgetting, we train with SGD on the non-IID datasets $\mathcal{D}_{t}$ as they appear in sequence. This method serves as an empirical lower bound, which any other model designed for federated reconnaissance should be able to beat. The Lower method updates the parameters of the model on the new mission classes in $\mathcal{D}_{t}$ without access to any previously seen data.
\item \textbf{iCaRL} We adapt the iCaRL~\citep{rebuffi2017icarl} algorithm for federated reconnaissance. The original iCaRL algorithm was developed for the single client scenario and has no inherent notion of model merging. We took inspirations from FederatedAveraging algorithm \cite{mcmahan2017communication} to make it work on the federated reconnaissance problem. Whenever each client goes to mission, it obtains an initial set of parameters from the server. During a mission, the client sees a number of examples from new classes it has not seen before and trains the model on those new classes and some examples from base classes that are stored in a set of exemplars stored by the iCaRL model. To resist catastrophic forgetting, the model is trained with a distillation loss on base exemplars and the cross entropy loss on exemplars from new classes. After the mission is complete, each client provides its new model and exemplars from new classes to the server. The server then gets models from multiple clients and merges them by averaging the common parameters from those clients. For each new class, a new neuron is added to the fully connected layer of the model and the weights are copied from the fully connected layer of the client model corresponding to the class being added. If more than one client learned the same class, the neuron parameters are averaged, again similarly to the FederatedAveraging algorithm. This averaged model serves as an initial model which is further fine-tuned on base exemplars and new classes exemplars using gradient descent on the server. This model initialization can be thought of as a meta-initialization similar to Reptile~\citep{nichol2018first} where each client update training can be regarded as the task specific training and the model merging can be thought of as an update in the Reptile model with a meta-learning rate equal to 1. This final model serves as an initialization for next client going on a mission.
\item \textbf{Federated Prototypical Networks} We extend prototypical networks for distributed, continual learning by deriving algorithms for online prototype updating and model merging. We describe this method in greater detail in section \ref{federatedprotonets}.
\item \textbf{Upper Baseline} To demonstrate the best case test-set results of a neural network architecture with access to all data seen by all clients and a liberal compute budget, during both on-client and on-server learning, we train with SGD on the joint distribution of all training data from all clients. This model serves as an empirical upper bound and comes at the storage and communication cost of saving and transferring all examples seen by all clients back to the server after each mission and retraining on all this data anytime new classes or examples are to be learned. Due to excessive computational expense, we did not evaluate the Upper baseline on the entire multi-mission \frb. Instead, we evaluated the model once by training with all training data of all classes, in the same way it would have been evaluated at the end of the benchmark.
\end{enumerate}

To put federated prototypical networks into the context of existing work, we also compare to a state of the art few-shot learning method Online aware Meta-learning (OML) from ~\citet{javed2019meta} on existing, single client, few-shot continual learning benchmarks. We include additional details on the above learning algorithms and evaluation procedures including hyperparameters in the supplementary material.

\subsection{Federated Prototypical Networks}
\label{federatedprotonets}
We propose to use prototypical networks to efficiently learn new classes in sequence. Given that prototypical networks are not gradient-based at test-time, they can be made to be robust to catastrophic forgetting when learning new classes by discriminative pretraining on a sufficiently large number of classes. When evaluated on a federated reconnaissance benchmark, we can compute an unbiased estimate of the mean (and the variance if we so desire) for each class by simply storing the previous prototype (, variance,) and the number of examples used to compute the previous prototype. We define prototypical networks following ~\citep{snell2017prototypical}:
\begin{align}
	\mathbf{z} = f_{\theta}\left(\mathbf{x}_{i}\right) \\
	\mathbf{\bar{z}}_{j}=\frac{1}{\left|S_{j}\right|} \sum_{\left(\mathbf{x}_{i}, y_{i}\right) \in S_{j}} f_{\theta}\left(\mathbf{x}_{i}\right)
\end{align}
in which $f$ is a neural network embedding function parameterized by $\theta$ and $S_j$ is the support set of the class $j$. Training a prototypical network proceeds be minimizing the cross entropy loss over query examples, where the predicted class is taken as the softmax over negative Euclidean distances $d(\cdot)$ between query embeddings and the prototypes of the support data:
\begin{equation}
\label{eq:proto-prob}
	p_{\theta}(y=j \mid \mathbf{x})=\frac{\exp \left(-d\left(f_{\theta}(\mathbf{x}), \mathbf{\bar{z}}_{j}\right)\right)}{\sum_{j^{\prime} \in J} \exp \left(-d\left(f_{\theta}(\mathbf{x}), \mathbf{\bar{z}}_{j^{\prime}}\right)\right)}
\end{equation}

Now we would like to be able to compute unbiased estimates of prototypes for classes that are observed by multiple clients at the current time step or have been observed previously in exposure history. To improve storage and communication efficiency, instead of storing all raw examples for a class or even all example embeddings for a class,
we instead can compute an unbiased running average for each prototype by storing the previous prototype and the number of examples used to compute it:
\begin{equation}
	\mathbf{\mu}_t = \frac{k_{t - 1} \mu_{t - 1}}{k_t} + \frac{(k_t - k_{t - 1}) \mathbf{\bar{z}}_{j}}{k_t}
\end{equation}
where $k_t$ is the number of examples in total observed of class $j$ at time $t$, $k_{t -1}$ is the number of total classes observed at time $t -1$ for class $j$, $\mathbf{\bar{z}}_{j}$ is the centroid for class $j$ at time $t$, and $\mathbf{\mu}_t$ is the online average of $z$ from all examples for class $j$.
Modulo numerical issues, via the law of large numbers, such a running average will converge to the true prototype $\mu^{*}$ for each class, given a fixed parameterization of the embedding function $f_{\theta}$.
\begin{equation}
	\bar{\mathbf{z}}_{k} \stackrel{\text { a.s. }}{\longrightarrow} \mathbf{\mu^{*}} \quad \text { as } k \rightarrow \infty
\end{equation}
Numerical issues cannot be so easily ignored in practice, so we use a more numerically stable algorithm for online averaging as proposed by~\citet{west1979updating, schubert2018numerically}:
\begin{equation}
	\mathbf{\mu}_t \leftarrow \mathbf{\mu}_{t - 1} + \frac{k_t - k_{t - 1}}{k_t} \left( \mathbf{\bar{z}} - \mathbf{\mu}_{t - 1} \right)
\end{equation}

Putting these ideas together, we arrive at Algorithm \ref{alg:fedprotonetsonline}, which implements a mulit-client learning and model merging routine with prototypical networks. Algorithm \ref{alg:fedprotonetsonline} can be used for on client learning and knowledge transfer with a central server and even other clients for fully decentralized peer-to-peer learning. In addition to the algorithm shown in \ref{alg:fedprotonetsonline}, we also evaluate a variant of the algorithm which removes the numerical effects of computing online averages by simply storing and transferring embeddings of all examples seen and computing the prototypes only at inference time.
\begin{algorithm}[tb]
   \caption{Federated Prototypical Networks: Online Averaging \& Model Merging}
   \label{alg:fedprotonetsonline}
\begin{algorithmic}
   \STATE {\bfseries Input:} $k$-shot dataset $\mathcal{D}$ of $n$ classes, previous prototypes $\mathbf{\bar{z}}_j$, previous class counts $k_{j, t-1}$
   \FOR{client $c_i$  {\bfseries in} $\mathbb{C}$}
    \STATE $\mathbf{\bar{z}}_{j, i} \leftarrow \frac{1}{k_{j, i}}\sum_{x, y}f_{\theta}(x_i)$ for $(x_j, y_j) \in \mathcal{D}$
   \ENDFOR\\
   \COMMENT{On-client model is optionally evaluated in situ.}\\
   \COMMENT{Merge prototypes to server:}
   \FOR{client $c_i$  {\bfseries in} $\mathbb{C}$}
    \FOR{class $j \in \mathcal{D}$}
    \STATE $k_{j, t} \leftarrow k_{j, t-1} + k_{j, i}$
    \STATE $\mathbf{\mu}_t \leftarrow \mathbf{\mu}_{t - 1} + \frac{k_t - k_{t - 1}}{k_t} \left( \mathbf{\bar{z}} - \mathbf{\mu}_{t - 1} \right)$
    \ENDFOR
   \ENDFOR
   \STATE Return all centroids $\{\mathbf{\bar{z}}_{j} | j \in \mathbb{M}_t \}$ back to clients
\end{algorithmic}
\end{algorithm}

For our experiments with prototypical networks, we reproduced the model architectures used in the original paper~\citep{snell2017prototypical}, though also experimented with a few key additional features. First, we find that overfitting to the pretraining base classes is a significant problem in the small-scale \mi \fr~benchmark we propose, so additional regularization is necessary. We find that applying dropout with a droprate of 0.2 in the embedding space significantly improves learning new classes. Furthermore, we find that the accuracy performance after ingesting new examples from classes quickly plateaus for prototypical networks when ingesting more examples than they were trained on. The problem of prototypical networks over-specializing the pre-trained model to solving $k$-shot problems, where $k$ is equal to the number of training shots per class, has also been noted in recent literature by \citet{triantafillou2020learning}. To address this problem, we find that the simple, model agnostic augmentation strategy of sampling $k$ at each iteration during pretraining yields significant benefits. We call this procedure $k$-shot augmentation and find that uniform sampling $k$ from $[5, 50]$ during pretraining significantly improves distributed, continual learning.

While in this work we assume that each client communicates directly to a central server, the extension of direct peer-to-peer distributed learning of federated prototypical networks is straightforward, assuming that all examples observed by all clients are unique. In an environment in which all examples are not unique, such as is common in social media with reshared images, additional bookkeeping would be required to avoid double counting of embeddings and to keep the estimate of the prototype unbiased.

\subsection{Neural Network Architectures}  

We evaluate all learning algorithms with two different neural network backbones. First, in line with a large body of work on meta-learning, we use the typical 4-convolutional layer model (denoted 4conv in figures) as used in~\citet{snell2017prototypical, finn2017model} and many other works. This model contains 4 layers each with $3\times3$ 2-d convolution with 64 output channels, batch normalization, a reLU non-linearity, and finally $2\times2$ maxpooling to cut the spatial dimensions in half at each layer. For the prototypical network model, we unroll the final feature channels following~\citet{snell2017prototypical}\footnote{We also experimented with average pooling the feature maps into a 64-element vector, though did not find that this significantly changed results from unrolling the feature maps into 1600 element vector. Furthermore, we found that we could reduce the final layer convolutional output channels to 32 element feature vectors without loss of accuracy, which identifies an additional means of compression when bandwidth and storage efficiency are critical.}. In line with more recent work~\citet{triantafillou2020meta}, we also evaluate all learning algorithms with a resnet-18 backbone \citep{he2016deep}, which is a variant of the common residual network model that contains 18 layers.

\section{Evaluation}
\subsection{Single Client Continual Learning}
To put the strengths of using prototypical networks for continual learning into the context of prior work, we evaluate a single client continual learning benchmark put forth by~\citet{javed2019meta} in which a learner is exposed to 30 examples from classes seen in non-overlapping succession. On this evaluation, we compare prototypical networks to the OML method proposed by ~\citet{javed2019meta} in a local reproduction. Following ~\citet{javed2019meta}, the accuracy across all classes seen to date is computed at each evaluation.

\subsection{\mi Federated Reconnaissance Benchmark}
To evaluate federated reconnaissance, a new benchmark was required. Taking inspiration from~\citet{javed2019meta}, we adapted the popular \mi \citep{vinyals2016matching}~dataset in order to create the \mi Federated Reconnaissance Benchmark. The \mi dataset is composed of 100 classes taken from the Imagenet large scale visual recognition challenge~\citep{russakovsky2015imagenet}, with 600 examples each. We split the classes in half, yielding 50 base classes for pretraining and 50 field classes for online learning. Base classes are not seen again during online learning. The examples for each class are split into 500 training examples and 100 test examples for each class. In the default parameterization of the benchmark, we instantiate 5 clients and, during online learning, for each mission we sample 5 classes per client from field classes and 30 images per class. We resize all images to $84\times84$. We sample examples without replacement and the evaluation ends when all training examples have been sampled. To simplify the notation and the benchmark's implementation, each client learns classes on its mission in serial and the server model is updated synchronously. In practice, client learning will happen in parallel and the server model can be updated asynchronously.

The problem with the simple accuracy metric in \ref{eq:acc-eval-2} is that in the presence of a pretraining dataset the metric will bias towards the pretraining classes in $\mathbb{B}$ until a sufficient number of new classes have been observed. To summarize the learning dynamics of federated reconnaissance, a metric that balances across base and field classes is required. We simply weight these two accuracies computed on the test sets of classes equally for all results reported in \ref{results}:
\begin{equation}
\label{eq:acc-avg}
acc_{avg} = \frac{acc_{base} + acc_{field}}{2}
\end{equation}
In \ref{results}, we report results from evaluating this accuracy metric after models have been merged back to the server from clients (Eval 2 diamond in figure \ref{fig:fedreconeval}). All results are evaluated on the union of the set of classes that any client has seen thus far at mission $t$, $\mathbb{M}_t$. All accuracy metrics we report here are computed against the test set $p_{test}(y|x)$ for both base and field classes.

\section{Results}
\label{results}
\subsection{Single Client Continual Learning}
\begin{figure}[h]
	\centering
	\includegraphics[width=.75\linewidth]{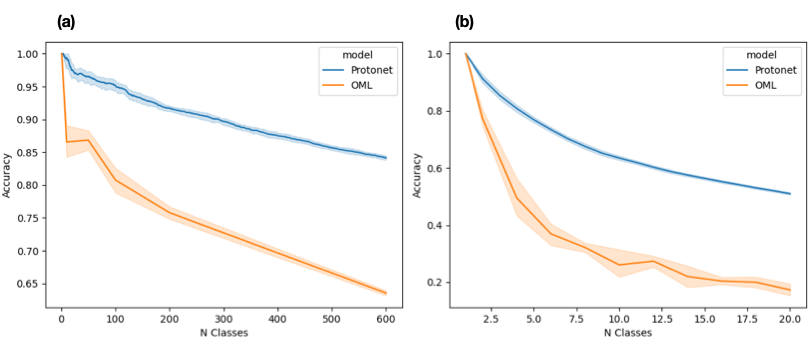}  
	\caption{Single-client test-set continual learning results on Omniglot (a) and \mi~{b}. Accuracy is evaluated on test examples from tasks not seen during pretraining. At each evaluation step, the model is trained on only the classes added at that step and then is evaluated on held out examples from all previously seen classes.}
	\label{fig:fig1}
\end{figure}

We first present results of evaluating our implementation of prototypical networks on the single client variant of federated reconnaissance. Specifically, we reproduce the benchmark used in~\citet{javed2019meta, beaulieu2020learning} and find that a simple four convolutional layer prototypical network outperforms the larger more complex model and learning algorithm of ~\citet{javed2019meta} that was specifically designed for continual learning. Prototypical networks increase the accuracy by over 22\% (absolute percentage points) after learning 600 Omniglot classes and over 33\% (absolute percentage points) after learning 20 mini-ImageNet classes incrementally, as shown in Figure \ref{fig:fig1}.


\subsection{Federated Reconnaissance}
\begin{table}[t]
\caption{Accuracies and computational complexity of learning algorithms for federated reconnaissance. Accuracy is evaluated at the end of federated reconnaissance on a 100-way classification problem. All experiments were repeated five times and accuracy values show mean and 95\% confidence intervals. Evaluation time
is shown for one run of the \frb. Asymptotic analysis describes one mission of learning new data where $E$ is the number of training epochs for gradient-based methods, $J$ is the total number of classes seen thus far, $K$ is the total number of examples per class seen thus far, $\mathcal{D}$ is the few-shot new dataset, $\gamma$ is the compression factor of the embeddings (which, while constant, we include to differentiate from image space)}, and $\mathtt{Buffer}$ is the fixed-size iCaRL buffer for exemplars.
\label{sample-table}
\vskip 0.15in
\begin{center}
\begin{small}
\begin{tabular}{lcccr}
\toprule
Learning  & $Acc_{avg}$ & Eval & Time  & Space \\
alg.      &             & time & cmplx & cmplx \\
\midrule
Upper    & $47.1 \pm 0.0\%$& 4days & $\mathcal{O}(EJK)$ & $\mathcal{O}(JK)$ \\
k-aug. protos & $37.4 \pm 0.0\%$ & 1.5hrs & $\mathcal{O}(|\mathcal{D}|)$& $\mathcal{O}(\frac{JK}{\gamma})$ \\
k-aug. onl. protos & $36.3 \pm 0.1\%$ &  1.5hrs & $\mathcal{O}(|\mathcal{D}|)$ & $\mathcal{O}(\frac{J}{\gamma})$ \\
Protonets & $34.2 \pm 0.0\%$ &  1.5hrs& $\mathcal{O}(|\mathcal{D}|)$ & $\mathcal{O}(\frac{JK}{\gamma})$ \\
Onl. protos & $33.6 \pm 0.1\%$ &  1.5hrs& $\mathcal{O}(|\mathcal{D}|)$ & $\mathcal{O}(\frac{J}{\gamma})$\\
iCaRL & $20.1 \pm 1.7\%$ & 4.5hrs & $\mathcal{O}(E)$ & $\mathcal{O}(\mathtt{Buffer})$\\
Lower & $1.3 \pm 0.3\%$ & 3hrs & $\mathcal{O}(E|\mathcal{D}|)$ & $\mathcal{O}(1)$\\
\bottomrule
\end{tabular}
\end{small}
\end{center}
\vskip -0.1in
\end{table}

\begin{figure*}[h]
	\centering
	\includegraphics[width=\linewidth]{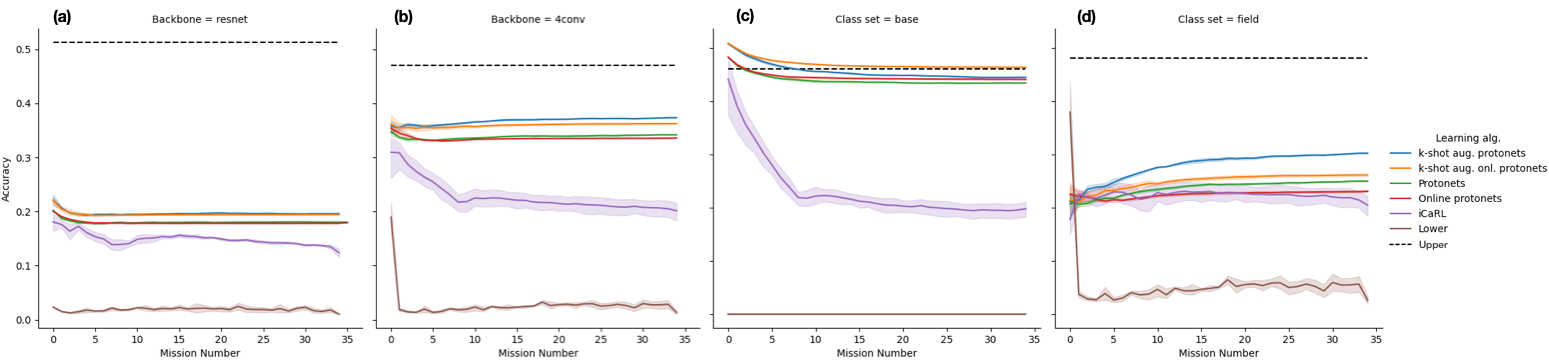} 
	\caption{(a) and (b) plot federated reconnaissance results partitioned by neural network backbone and learning algorithm. Accuracy is evaluated on the test set for each class and averaged according to eq. \ref{eq:acc-avg} to balance base and field classes equally throughout evaluation. (c) and (d) plot federated reconnaissance results for the four layer convolutional model with the accuracy components of eq. \ref{eq:acc-avg} partitioned into base classes (c) and field classes (d). Higher accuracy on base classes indicates resistance to catastrophic forgetting while higher accuracy on field classes indicates ability to learn new classes presented in non-IID data online. The results of training the Upper model once with all training data of all classes, in the same way it would have been evaluated at the end of the benchmark is shown in the black dashed line for reference.}
	\label{fig:fed-recon-eval}
\end{figure*}


\begin{figure}[h]
	\centering
	\includegraphics[width=0.35\linewidth]{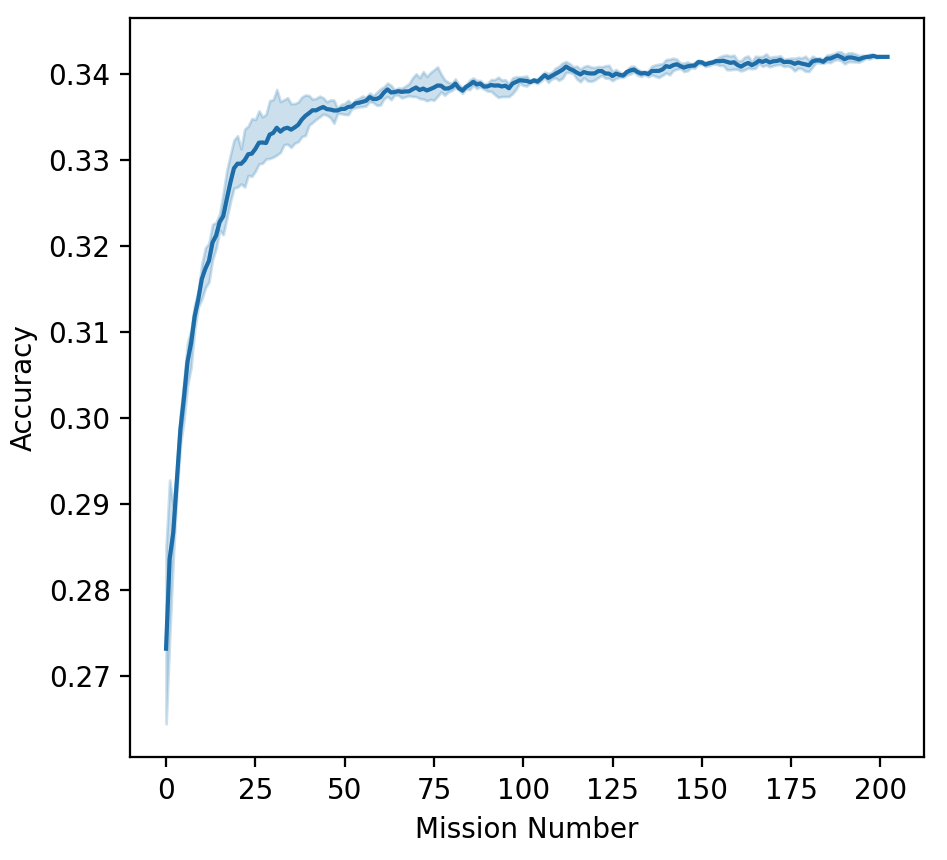} 
	\caption{Collective ascent of accuracy after model merging (Eval 2 diamond in \ref{fig:fedreconeval}) of federated prototypical networks as multiple clients gather and share information via algorithm \ref{alg:fedprotonetsonline}. In this experiment, each client receives 5 examples per class on each mission, instead of 30.  Accuracy is evaluated following eq. \ref{eq:acc-avg} on the \mi~\frb~on all classes that any client has seen thus far, including base classes. Recall that during \fr~evaluation, examples are only seen once from field classes. This means that an accuracy improvement requires that the average of backward transfer (accuracy on previously seen classes) and forward transfer (accuracy on new classes) must be positive.}
	\label{fig:collective-ascent}
\end{figure}

Of all learning algorithms evaluated on the  \mi~\frb, we find that federated prototypical network variants are the most accurate and computationally efficient\footnote{Recall that due to excessive computational expense, we did not evaluate the Upper baseline on the entire multi-mission \frb.}. In particular, we find that using $k$-shot augmentation during pretraining of prototypical networks is an effective means of improving the \frb~accuracy of prototypical networks across a range of $k$ values. When comparing storing all embeddings in memory to computing the prototypes via online averaging, we find a slight, though statistically significant, degradation in accuracy due to numerical issues. We believe these numerical issues are due either to floating point imprecision, inherent non-determinism, catastrophic cancellation, or, likely, a combination of some number of those. We leave root causing these issues and the development or more numerically robust online prototype computation to future work.
It is also clear from the results shown in figure \ref{fig:fed-recon-eval} that the 4-layer convolutional model substantially outperforms the resnet-18 model across learning algorithms except for the upper baseline. The resnet model has significantly more capacity than the 4-layer convolutional model and we find that it overfits excessively to the pretraining examples. For these reasons, we focus further more in-depth analyses onto the results of the 4 layer convolutional model.

To better understand the learning dynamics of the algorithms evaluated on the \frb, we decompose $acc_{avg}$ shown in eq. \ref{eq:acc-avg} into its constituent base and field accuracies as shown in figure \ref{fig:fed-recon-eval} c and d. We find that prototypical networks are not only able to resist catastrophic forgetting of base classes and examples seen earlier during the \frb~but can do so while improving accuracy on field classes. These results are in stark contrast to our adapted version of iCaRL, which suffers from catastrophic forgetting while also being unable to improve its accuracy in distinguishing new concepts as more data becomes available.

Furthermore, we are able to demonstrate collective ascent with prototypical networks by showing that a set of clients can gather data in parallel and communicate sparsely while still improving their accuracy on a shared knowledge base comprising a growing set of classes even without observing the examples of those classes directly. In figure \ref{fig:collective-ascent}, each client receives 5 examples per class on each mission instead of 30. Such a parameterization of the experiment causes the difficulty (i.e. the ``way'') of the problem to increase more rapidly relative to the amount of data seen for each class, making it clear that the group of learners is effectively leveraging shared knowledge.

\section{Discussion}
In this work, we have presented federated reconnaissance, a new a class of learning problems in which distributed clients learn new concepts independently and must be able to communicate that knowledge efficiently. We proposed an evaluation framework and  evaluated a number of baseline learning algorithms for distributed continual learning. In particular, we derive simple algorithms for efficiently leveraging prototypical networks and find that they are a strong baseline method for \fr~ and class-incremental continual learning. These results suggest that the simple idea of passing feature vectors is an important avenue for future research on \fr~ and continual learning more generally. In future work, we plan to extend this research to openset problems and out-of-distribution class identification on larger benchmarks.


%
%
%


\section{Acknowledgements}
This work was partially funded by U.S. Naval Sea Systems Command (NAVSEA) through the Small Business Technology Transfer (STTR) program, grant number N68335-20-C-0788. The authors would like to thank Andrew B. Leach for his excellent and discerning feedback on an earlier version of this manuscript.

\bibliographystyle{unsrtnat}






\newpage
\begin{appendices}

%
%

%

\section{Introduction}
We present supplementary material for our paper Federated Reconnaissance: Efficient, Distributed, Class-Incremental Learning. This supplementary material is mainly aimed at describing experimental details regarding reproducibility and hardware though also includes an additional ablation to inspect the effects of the number of base training classes.

\section{Experiments}
All experiments were run 5 times to produce mean and confidence intervals of test-set metrics. Evaluation times are given in the main body of the paper.
\subsection{Online Gradient-Based Methods}
For pretraining the iCaRL, Lower Baseline, and Upper Baseline models, we pretrain on base 50 classes as described in the main body of the paper on the evaluation framework for the \mi~\frb except for upper method which starts with a model trained on all 100 classes. We trained the resnet-18 and the 4 layer convolutional network backbones using stochastic gradient descent up to 200 epochs with an initial learning rate 0.01, weight decay 1e-2, Nesterov momentum 0.9, and step learning rate decay with step size 50 and gamma 0.1. Validation dataset from 50 base classes is used to select the best model. The hyper-parameters for the online updates during \fr~evaluation are chosen by training the base model on 40 base classes and updating the model on a hold out set of 10 evaluation classes.

In greater detail for these methods which use gradients for learning new classes and examples during \fr~evaluation:
\begin{itemize}
    \item \textbf{iCaRL}:\\
    During a mission, each client updates the model with a stochastic gradient descent and a learning rate 0.01, Nesterov momentum 0.9 up to 30 epochs. The temperature parameter for the distillation loss is set to 2. We used a budget of 4000 total examples. Examples are chosen at random to maintain in the cache, balancing across classes, meaning that there are 40 examples for each of the 100 \mi~classes that are stored, transferred between client and server, and used for training. After the model is merged, 50 steps of stochastic gradient descent are carried out on base exemplars and new class exemplars with a learning rate 0.01, weight decay 1e-3, Nesterov momentum 0.9, and step learning rate decay with step size 30, and gamma 0.1.

    \item \textbf{Lower:}\\
    During a mission, each client updates the model on new classes with a stochastic gradient descent with learning rate 0.01, Nesterov momentum 0.9, and a step learning rate decay with a step size of 30 and gamma of 0.1 up to 50 epochs.

    \item \textbf{Upper:}\\
    For the upper baseline we only evaluated the learning dynamics at the end of the \frb, due to excessive computational costs. We trained the two backbones on the training dataset of all 100 classes for 200 epochs using stochastic gradient descent with a learning rate 0.01, weight decay 1e-2, Nesterov momentum 0.9, step learning rate decay with step size 50 and gamma 0.1.
 \end{itemize}

\subsection{Federated Prototypical Networks}
For pretraining the prototypical network backbones on the base classes, we closely follow the original implementation of~\citet{snell2017prototypical}. As such, we pretrain using the Adam optimizer \citep{kingma2015adam} for 30,000 episodes (i.e. gradient steps) with an episode/batch size of 5 randomly sampled classes with 5 support and 5 query examples per class. We use an initial learning rate of $1e-3$ which we half every 2000 episodes. The only significant changes we make to the original training regime are directed at reducing overfitting the training episodes. Namely, we:
\begin{enumerate}
  \item Train with dropout \citep{srivastava2014dropout} in the embedding space, with a drop probability of 0.2, and
  \item Implement a novel augmentation procedure, which we call $k$-shot augmentation, in which we sample the number of support and query examples from a uniform distribution over $[5, 50]$.
\end{enumerate}

\subsection{Ablation of number of base training classes}
\begin{figure*}[h]
	\centering
	\includegraphics[width=0.35\linewidth]{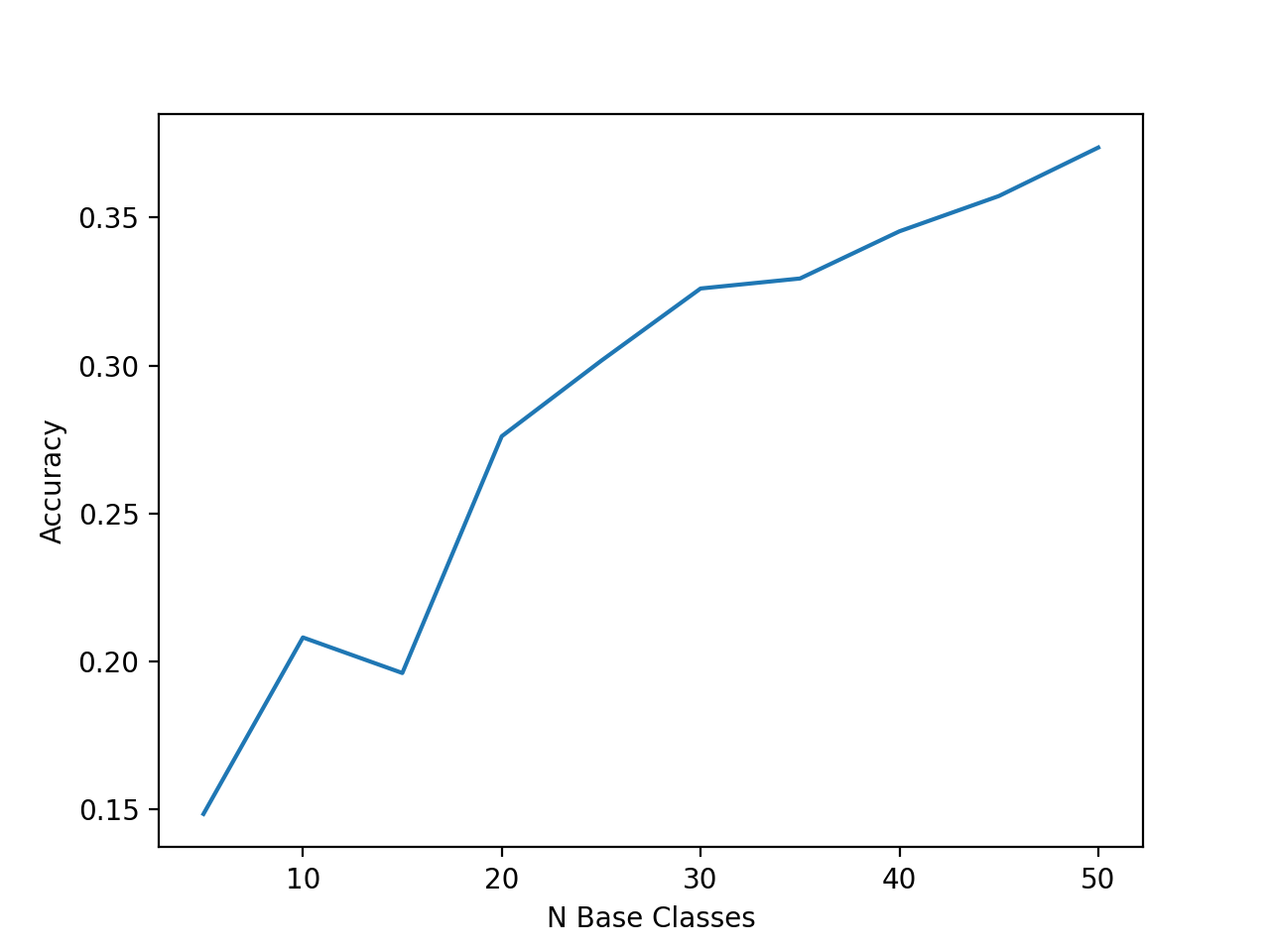}
	\caption{The final averaged accuracy $acc_{avg}$ on \frb for the k-shot augmented prototypical networks when pretrained with across a varying number of base classes.}
	\label{fig:acc-vs-n-base-classes}
\end{figure*}

To inspect the effects of the number of base classes on the final $acc_{avg}$ of the \mi \frb, we took the best learning algorithm and model architecture, the k-shot augmented prototypical network, and trained it on only a random selected subset of base classes. After this, we ran the rest of the benchmark as usual, learning an additional 50 classes in a distributed fashion. Due to computational expense, we only ran this experiment once per number of base classes. We found that the final $acc_{avg}$ at the end of the \frb increased nearly linearly with respect to the number of base classes as shown in Figure \ref{fig:acc-vs-n-base-classes}. This highlights the importance of a large number of base classes for class incremental learning with embedding models.

\section{Hardware}
All experiments were run on a single GPU. We mock the parallel learning of clients for the purposes of initial experimentation by running each client in serial. Most experiments were run on NVIDIA P100 GPU nodes with 28 CPU cores and 224 GB of memory, from which we report evaluation times. A small number of additional experiments were run on an NVIDIA RTX 2080 GPU.

%
%

\end{appendices}

\end{document}